\documentclass{article}

\PassOptionsToPackage{numbers,compress}{natbib}
\usepackage[preprint]{tackling_climate_workshop_style}

\usepackage[utf8]{inputenc} 
\usepackage[T1]{fontenc}    
\usepackage{hyperref}       
\usepackage{url}            
\usepackage{booktabs}       
\usepackage{amsfonts}       
\usepackage{nicefrac}       
\usepackage{microtype}      
\PassOptionsToPackage{hyphens}{url}
\usepackage[utf8]{inputenc}
\usepackage[T1]{fontenc}
\usepackage{booktabs}
\usepackage{graphicx}
\usepackage{float}
\usepackage{amsmath,amssymb}
\usepackage{microtype}
\usepackage{url}
\usepackage{xcolor}
\usepackage{subcaption}
\usepackage{hyperref}
\usepackage{cleveref}
\title{Closing the Loop on Contrail Avoidance\\
with Satellite Verification}

\author{%
  Spandan Ghose Chowdhury\\
  School of Computing\\
  Georgia Institute of Technology\\
  Atlanta, GA, USA \\
  \texttt{spandan\_gc@gatech.edu} \\
}

\begin{document}
\maketitle

\begin{abstract}
Contrails are the thin ice clouds that aircraft leave behind. They cause a large share of aviation's warming, and rerouting the few flights that produce them could avoid much of it. However, an avoided contrail only counts if a satellite can confirm that it never formed, and this check is hard: contrails are one to two pixels wide, cover only $0.18\%$ of pixels, and look very similar to natural cirrus. We build a small diffusion model (8.4M parameters, trained on one GPU) that detects them, and we run a controlled study to find out which components matter. The model reaches $0.476$ PR-AUC, compared with $0.414$ for a DeepLabV3+ baseline and $0.119$ for an adapted MedSegDiff. Doubling the input resolution of the CNN brings it to parity ($0.499$, $p{=}0.07$). Three lessons apply beyond contrails. First, check the input resolution before designing a new architecture. Second, simple flips and rotations more than double accuracy and matter more than any architectural choice we measured. Third, pretraining the model on contrail shapes is harmful: the model learns that thin strokes appear everywhere and paints them onto empty scenes. Precision collapses to $1\%$ while recall-based metrics still rate the degraded model as excellent, and no threshold or guidance heuristic repairs this failure.
\end{abstract}

\section{Introduction}

Aviation warms the climate in two ways: by emitting CO$_2$ and by producing condensation trails. Contrails that persist spread into cirrus-like sheets that trap outgoing heat, and contrail cirrus is assessed as the largest single component of aviation's effective radiative forcing, although the uncertainty is wide \citep{lee2021aviation,karcher2018formation,teoh2024global}. About $2\%$ of flights cause $80\%$ of annual contrail energy forcing \citep{teoh2024global}, and small changes in route or altitude let an aircraft avoid the ice-supersaturated layers in which contrails form \citep{teoh2020mitigating}. Unlike CO$_2$ warming, this warming can therefore be reduced quickly and at low cost.

In August 2026 the UK government, NATS, the Met Office, Google and academic partners launched Operation Blue Skies, a \pounds5M programme that applies AI-recommended altitude changes in Shanwick oceanic airspace, which accounts for about $5\%$ of global contrail warming, over the winters of 2026--27 and 2027--28 \citep{blueskies2026,googleblueskies2026}. A programme of this scale is only as credible as the measurement that follows it, and that measurement comes from satellite imagery: to claim that a contrail was avoided, someone must check whether it formed \citep{geraedts2024scalable}. This measurement step, rather than forecasting or routing, is the focus of this paper.

\begin{figure}[h]
  \centering
  \includegraphics[width=0.63\linewidth]{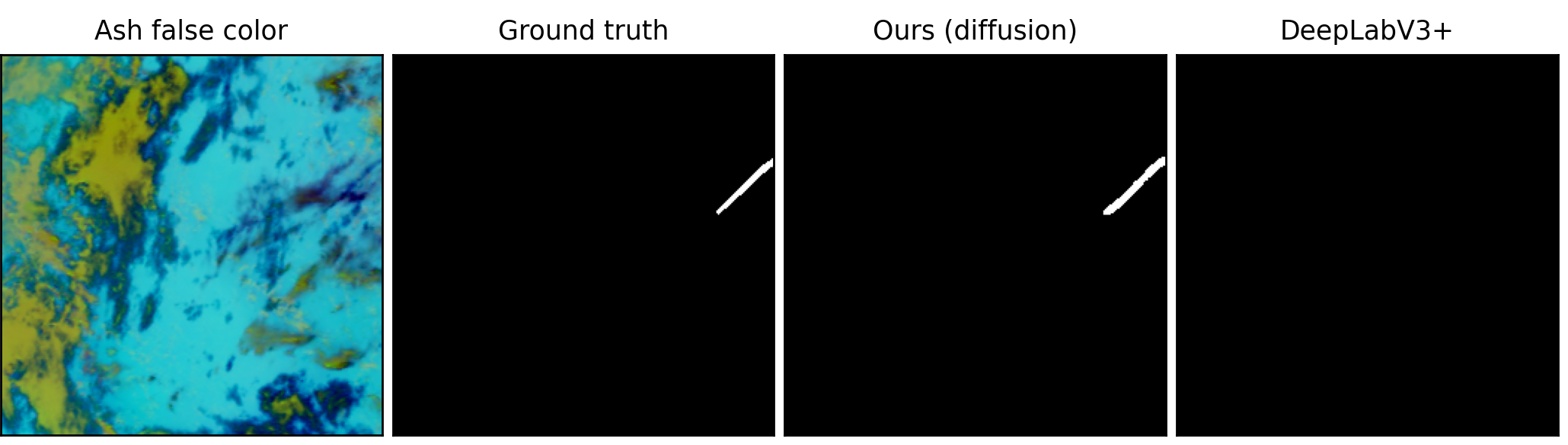}
  \caption{Spot the contrail: our model recovers a labeled contrail that is
  nearly invisible in GOES-16 infrared imagery; the DeepLabV3+ baseline misses
  it entirely. A contrail that is never detected is avoided warming that is
  never credited.}
  \label{fig:teaser}
\end{figure}

OpenContrails \citep{ng2024contrail} turned this task into a public benchmark with about 22{,}000 human-labeled scenes from GOES-16 imagery \citep{schmit2017abi}. Three properties make the task difficult. The targets are \emph{thin}: one or two pixels wide, so a small positional error destroys the overlap with the label. They are \emph{sparse}: $0.18\%$ of test pixels are positive and $70\%$ of images are empty. And they are nearly \emph{indistinguishable} from the hardest negatives, because thin cirrus shares their shape, brightness and texture (\cref{fig:teaser}). We therefore try a generative approach. A diffusion network \citep{ho2020denoising} builds its answer by repeated refinement rather than by a single forward pass, so it can return to a faint line instead of averaging it away. Running it several times also produces several masks, and their disagreement is a usable confidence signal. Diffusion has been applied to segmentation \citep{amit2021segdiff} and to thin tubular medical targets \citep{wu2023medsegdiff}, but whether its advantages survive a task this imbalanced had not been tested.

We answer this question in two parts. The first is a conditional diffusion network built for thin, linear structures. The second is a controlled study in which one protocol is frozen and every variant is retrained under it, including controls that give a conventional CNN the same test-time compute. Both parts are deliberately small (8.4M parameters against 26.7M for the baseline; 90 epochs on one GPU). This constraint is intentional: a verification pipeline that a weather service or a university group has to run cannot assume a large training budget. Two unexpected results are the part that transfers to other problems. Flip and rotation augmentation was the single most important ingredient, outweighing every architectural component we measured. Unconditional mask pretraining was actively harmful in a way that recall metrics do not reveal, and no threshold, calibration or classifier-free guidance \citep{ho2022classifier} repairs it.

\section{Related work}

\paragraph{Contrail detection from satellites.}
Automatic contrail detection from satellite imagery has a long history. The classical approach is the linear-contrail detection algorithm of \citet{mannstein1999operational}, which applies oriented line filters to AVHRR brightness-temperature differences; it was later extended to geostationary imagers to follow contrails through their life cycle \citep{vazquez2015contrail}. These hand-crafted detectors work well on bright, young contrails but miss faint or aged ones. Deep learning entered the field with \citet{kulik2019satellite} and \citet{meijer2022contrail}, who trained CNN segmenters on GOES-16 data and used them to estimate contrail coverage over the United States. \citet{ng2024contrail} released OpenContrails, the first large human-labeled dataset, together with a ResNet-based segmentation model trained under a large compute budget, and \citet{geraedts2024scalable} built on it to attribute detected contrails to individual flights. A public Kaggle competition on the same data \citep{kaggle2023contrails} showed that U-Net-style models with upsampled inputs and heavy test-time ensembling score well, and \citet{chevallier2023linear} combined detection, tracking and flight matching on Meteosat imagery. All of these detectors are single-pass discriminative models that report one score per pixel. Our work differs in three ways. We use an iterative generative model, which gives a per-pixel uncertainty map from the disagreement between samples. We train under a small, fixed budget, so that every baseline can be retrained under identical conditions and compared fairly. And we report which training choices matter, including one that silently fails, rather than only a headline score.

\paragraph{Segmenting thin, elongated and sparse structures.}
Contrails share their geometry with retinal vessels, roads, cracks and power lines: they are thin, long, connected, and a tiny fraction of the image. Work on these targets shows that per-pixel losses are the wrong objective, because a one-pixel shift destroys the overlap while leaving the shape intact. Topology-aware losses address this directly. \citet{mosinska2018beyond} penalize differences between deep features of the prediction and the label, \citet{hu2019topology} match persistent-homology features, and clDice \citep{shit2021cldice} compares the skeletons of prediction and label so that connectivity, rather than area, is rewarded. We adopt clDice because it is cheap to compute and it targets the failure we see most often, broken lines. Focal loss \citep{lin2017focal} is the standard answer to extreme class imbalance. The straightness of contrails links them to line detection: deep Hough-transform priors \citep{lin2020deep,zhao2022deep} accumulate evidence along candidate lines and improve the detection of faint straight structures. Our soft-Hough encoder is a fixed, lightweight version of this idea. Finally, diffusion models have been used for general segmentation \citep{amit2021segdiff} and for medical targets such as vessels and tumours \citep{wu2023medsegdiff}, and \citet{wolleb2022diffusion} showed that repeated sampling yields an implicit ensemble whose variance is a useful uncertainty estimate. We adapt MedSegDiff's frequency-domain conditioning to be direction-aware and use the same sampling ensemble for uncertainty, but to our knowledge none of these methods had been tested on a target this sparse.

\section{Method}

The design follows three properties of the target. Contrails are \emph{thin},
so a network that downsamples the image aggressively loses them; we therefore
keep a decoder branch that never downsamples. Contrails are \emph{straight},
because aircraft fly close to great-circle paths; fixed directional
line-averaging filters (a ``soft Hough'' encoder) supply this prior, although
our ablation cannot separate its effect from noise ($+0.005$, $p{=}0.35$).
Contrails also \emph{move and grow} between frames while cirrus mostly does
not, so frame differences provide a motion cue.

We follow the standard diffusion recipe \citep{ho2020denoising}. The mask is
corrupted with Gaussian noise at a random level, and a network learns to
predict the noise that was added, conditioned on four infrared frames.
Conditioning features enter at every scale and are modulated by the noise
level \citep{perez2018film}. They come from the two encoders above, from an
image encoder borrowed from super-resolution \citep{wang2018esrgan}, and from
a direction-aware version of MedSegDiff's frequency filter
\citep{wu2023medsegdiff}. Three auxiliary losses act on the reconstructed
mask: focal loss \citep{lin2017focal} for the extreme imbalance; soft clDice
\citep{shit2021cldice}, which rewards thin predictions that stay connected and
punishes broken ones; and a distance term that penalizes predicted mass far
from any label. Flips and rotations up to $\pm30^\circ$ are applied jointly to
the image sequence and the mask. At test time we generate $N$ masks
\citep{song2021denoising} and keep the pixels on which at least $60\%$ of them
agree, discarding connected components smaller than 50 pixels. The mean of the
masks is a dense confidence map, and their disagreement is a per-pixel
uncertainty map.

\begin{figure}[t]
  \centering
  \begin{subfigure}[t]{0.4\linewidth}
    \includegraphics[width=\linewidth]{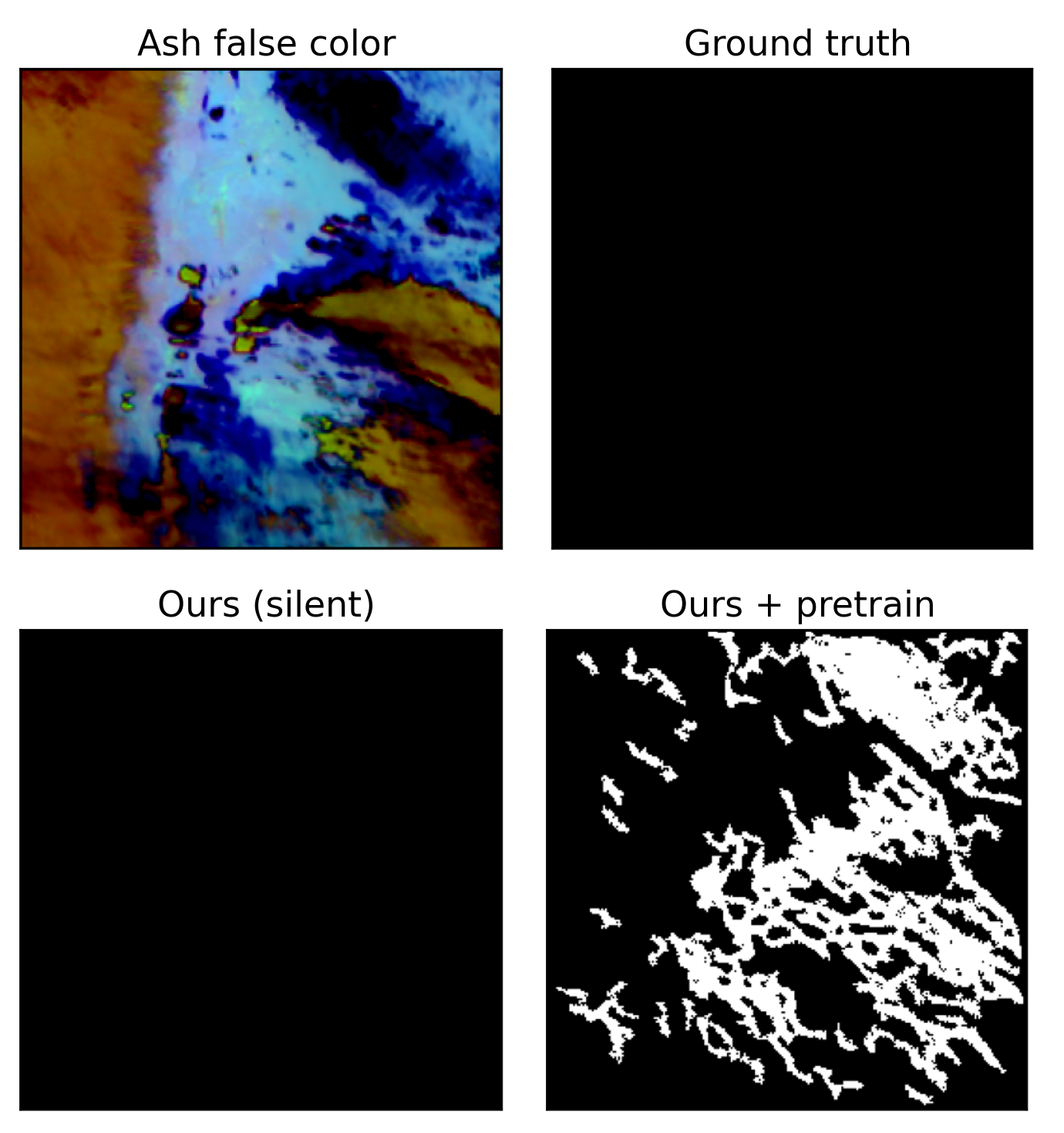}
    \caption{The pretraining trap.}
    \label{fig:pretrain}
  \end{subfigure}\hfill
  \begin{subfigure}[t]{0.6\linewidth}
    \includegraphics[width=\linewidth]{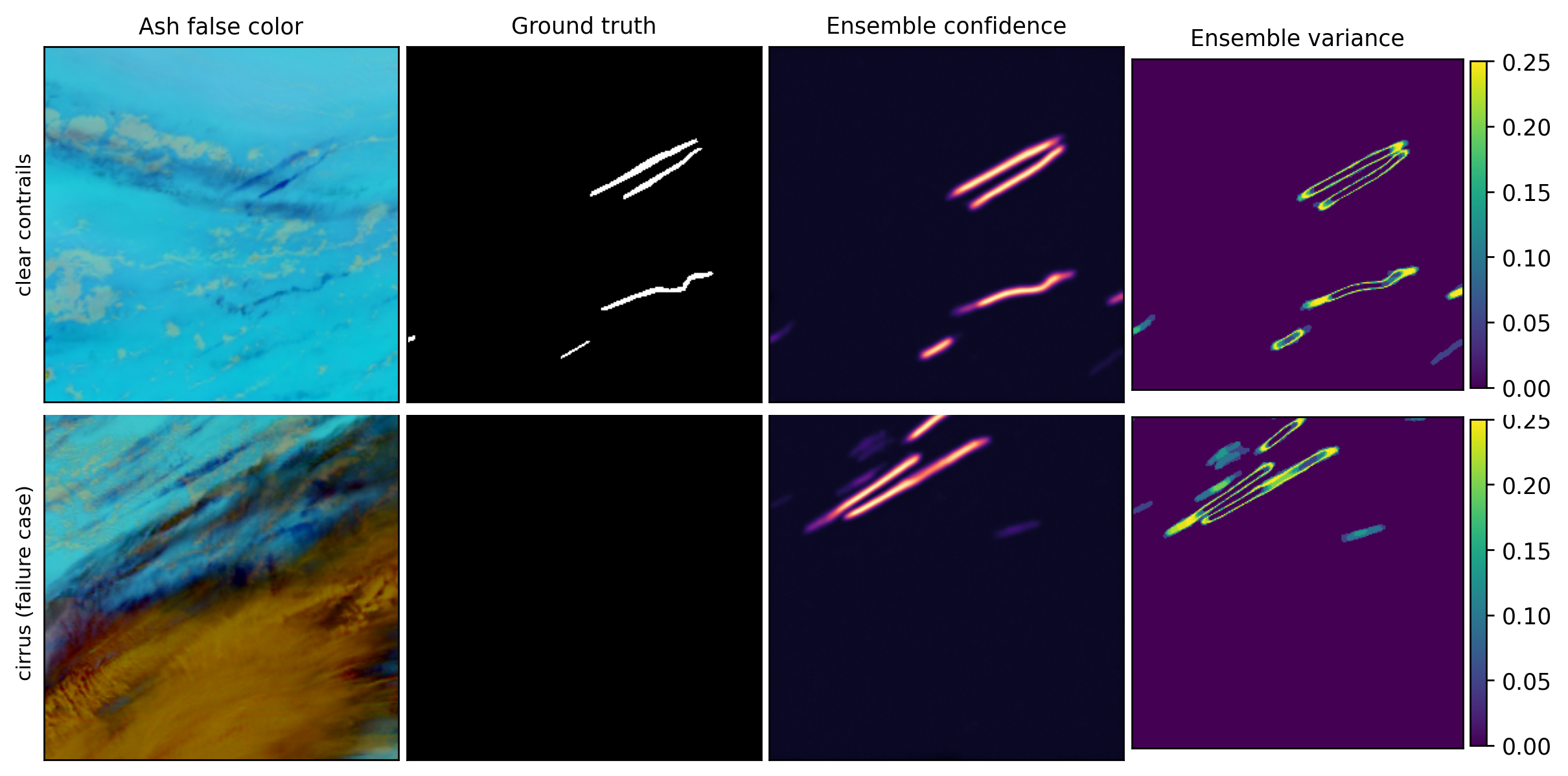}
    \caption{Ensemble uncertainty.}
    \label{fig:uncertainty}
  \end{subfigure}
  \caption{Two behaviours that matter for verification. (a)~Our recipe still
  fires on $42\%$ of empty scenes (compared with $2$--$3\%$ for DeepLabV3+);
  the scene shown is one it gets right. The unconditionally pretrained variant
  paints contrail-shaped strokes across the cirrus: $15.7\%$ of test pixels are
  marked positive against a prevalence of $0.18\%$, yet skeleton recall reads
  $0.96$. (b)~Input, label, ensemble confidence and vote variance. On real
  contrails (top row) the variance forms a thin outline around the detection.
  On cirrus false positives (bottom row) it thickens and stray patches appear
  away from the detections, which gives a signal for routing cases to review.}
  \label{fig:behaviours}
\end{figure}

\section{Experimental setup}

We use the full OpenContrails release \citep{ng2024contrail}: 20{,}529
training scenes and 1{,}856 validation scenes. Each scene is a $256\times256$
crop with four infrared frames and one human-labeled mask. Because the
official test set is not public, we split the validation scenes once, with a
fixed seed, into a val-tune half and a test half (928 scenes each) and froze
both halves before any experiment. Checkpoints and thresholds are chosen on
val-tune only. Every number below is computed once on the untouched test half,
with confidence intervals from a paired scene-level bootstrap.

We report three metrics. PR-AUC is computed over all test pixels. Skeleton
recall is the fraction of the label's centerline that the prediction covers;
it answers ``did we find it?''. HD$_{95}$ is a worst-case geometric error that
is driven by distant false-positive blobs. DeepLabV3+ \citep{chen2018encoder}
with an ImageNet-initialized ResNet-50 encoder \citep{he2016deep,deng2009imagenet}
and MedSegDiff \citep{wu2023medsegdiff} are both trained by us under identical
data, splits, budget, post-processing and metric code. The published
OpenContrails model is an external reference that was trained under a much
larger budget; the gap to it is not a labeling or prevalence artifact
(\cref{app:repro}).

\section{Results}

\paragraph{Primary comparison, and the control that reaches parity.}
\Cref{tab:main} gives the main finding: our model gains $+0.062$ PR-AUC over
DeepLabV3+ at a matched $256^2$ input ($95\%$ CI $[0.038,0.088]$) and reaches
four times the score of the adapted MedSegDiff. Our number comes from a
20-sample ensemble while the CNN's comes from a single pass, so we also give
the CNN the same test-time budget. Eight-fold test-time augmentation and a
five-seed deep ensemble \citep{lakshminarayanan2017simple} ($0.438$) each
close less than half of the gap. A \emph{discriminative twin}, which uses the
same encoders and losses but is trained as a one-shot segmenter with the mask
branch zeroed, falls below the plain CNN. The architecture alone therefore
does not explain the margin either. (The twin also loses the implicit
regularization of the diffusion objective, so we stop short of crediting
iterative sampling itself.)

One control does change the picture, and we report it rather than omit it.
Upsampling the input to $512^2$ lifts DeepLabV3+ to statistical parity with
our model. Resolution is a lever we have not yet given the diffusion model,
because its sampling cost grows with resolution. What survives across
resolutions is the recall side: our model has the best skeleton recall
($0.877$ against $0.836$ for the $512^2$ CNN) and recovers $62.0\%$ of labeled
contrail segments against $53.9\%$ (although the $512^2$ CNN has the better
segment F1). It also provides per-pixel uncertainty. The CNN keeps the best
worst-case geometric error, because our remaining failures are occasional
contrail-shaped false positives far from any label, and it is about
$4{,}900\times$ faster than our model at $N{=}20$.

\begin{table}[t]\centering
\caption{Frozen OpenContrails test split (928 images, prevalence $0.18\%$);
seed 42, two-seed mean $0.475\pm0.001$. The $512^2$ row upsamples the input
bilinearly and is scored on the $256^2$ grid; it is not statistically
separable from ours ($p{=}0.07$). Bold: best per column among compared rows.
Last row: external reference trained under a much larger budget, not a
within-protocol comparison.}
\label{tab:main}
\small
\begin{tabular}{@{}lccccccc@{}}\toprule
Method & Par. & PR-AUC$\uparrow$ & SkelRec$\uparrow$ & HD$_{95}\downarrow$ & P & R & F1 \\ \midrule
DeepLabV3+ & 26.7M & 0.414 & 0.816 & 57.3 & 0.420 & 0.520 & 0.465 \\
\;{+}8-fold TTA & 26.7M & 0.442 & 0.805 & 55.7 & 0.476 & 0.484 & 0.480 \\
\;$512^2$ input & 26.7M & \textbf{0.499} & 0.836 & \textbf{51.2} & \textbf{0.508} & 0.573 & \textbf{0.539} \\
MedSegDiff & 129M & 0.119 & 0.825 & 312.8 & 0.050 & 0.533 & 0.092 \\
Twin (ours, no diffusion) & \textbf{8.4M} & 0.209 & 0.781 & 113.8 & 0.352 & 0.309 & 0.329 \\
Ours & \textbf{8.4M} & 0.476 & \textbf{0.877} & 127.9 & 0.392 & \textbf{0.640} & 0.486 \\
\midrule
OpenContrails~\citep{ng2024contrail} & -- & 0.651 & -- & -- & -- & -- & -- \\
\bottomrule
\end{tabular}
\end{table}

\paragraph{Augmentation is the primary driver.}
A $2\times2$ factorial experiment separates the two training choices. Joint
flips and rotations raise the from-scratch model from $0.204$ to $0.424$
PR-AUC. With unconditional pretraining, the same two cells read $0.038$ and
$0.040$. (All four cells include a background-noise boost that the final
recipe drops; it is worth $0.052$. The pretraining arm keeps its own
fine-tuning rate, so the cells compare complete recipes.) Contrails occur at
every orientation, so rotation teaches the right invariance.

\paragraph{Unconditional pretraining is a pitfall.}
The idea sounds reasonable: first teach the denoiser what contrail masks look
like (mask only, no image), then fine-tune with conditioning. In both cells,
however, performance collapses to about $0.04$ PR-AUC, and the mechanism is
visible in \cref{fig:pretrain}. The model has learned that ``thin strokes exist
everywhere'' and paints them onto the $70\%$ of scenes that are empty, an
$87\times$ over-prediction. This is not a threshold problem (the best
achievable F1 is $0.094$), and skeleton recall stays at $0.96$, which is
exactly why the failure is dangerous. Classifier-free guidance
\citep{ho2022classifier} does not repair it: a guidance-trained pretrained
variant drops from $0.187$ PR-AUC without guidance to $0.029$ at $w{=}1.5$.
The shape prior must therefore come from augmentation and from the losses,
which carry the recipe: removing focal loss drops PR-AUC to $0.215$, and
removing clDice collapses precision from $0.39$ to $0.06$.

\section{Pathway to climate impact}

Contrail avoidance is a rare case of a large, fast and cheap climate lever
whose bottleneck is measurement rather than engineering. An operational
programme forecasts the ice-supersaturated regions, reroutes the few flights
that would cross them, and \emph{verifies} from satellite imagery whether the
contrail failed to appear. Only the verification step closes the loop.
Without trustworthy detection, a programme cannot tell a successful diversion
from a wrong forecast, and it cannot support the accounting that a mandate
would require. The users are concrete: evaluators of trials such as Blue
Skies, per-flight attribution pipelines \citep{geraedts2024scalable}, and
eventually a regulator asked to certify avoided warming.

\paragraph{Why these metrics.}
Because about $2\%$ of flights cause about $80\%$ of contrail forcing
\citep{teoh2024global}, verification is rare-event detection. Pixel accuracy
or IoU would be dominated by the $99.8\%$ of pixels that are empty. Recall
sets the ceiling: a contrail that is never detected is warming attributed to
nobody, and it cannot be recovered downstream. This is why we lead with PR-AUC
and per-contrail recall. Precision decides how much of a tally is real and how
much review it costs.

\paragraph{Where it could go wrong.}
A verifier that over-reports avoided contrails turns an avoidance programme
into a greenwashing instrument, and one that mis-attributes assigns the cost
to the wrong operator. The failure mode we document therefore matters more
than the headline number. Attribution filters false positives against flight
tracks, but hallucinated contrail-\emph{shaped} strokes are more likely than
blobs to land near a real track. This is why we foreground the pretraining
collapse over its flattering $0.96$ skeleton recall. The structured vote
variance (\cref{fig:uncertainty}) can route doubtful detections to review,
which turns our $42\%$ false-positive rate on empty scenes into a review
burden rather than a silent error.

\paragraph{Limitations.}
The case for this model rests on quality and uncertainty at $N\geq5$. It is
about $4{,}900\times$ slower than a CNN (still affordable at GOES-16's
10-minute cadence), less accurate at $N{=}1$, and matched on pixel PR-AUC by a
$512^2$-input DeepLabV3+ that we have not yet answered at that resolution. All
results are on one dataset and one sensor. Transfer to Meteosat, which covers
the airspace where Blue Skies flies, is the untested next step.

\paragraph{Acknowledgments.} This work began as a project in CS 7643 (Deep Learning) at Georgia Tech. I thank my project teammates, Daniel Milanes Perez, Mahsa Payami Shabestar, and Jordan Taylor, who introduced me to the contrail detection problem and and with whom I carried out the initial exploratory data analysis and some basic diffusion based models on OpenContrails. All modeling, architectures, experiments and writing in this paper are my own.

{\small
\bibliographystyle{unsrtnat}
\bibliography{main}

@article{lee2021aviation,
  author  = {Lee, David S. and Fahey, David W. and Skowron, Agnieszka and Allen, Myles R. and Burkhardt, Ulrike and Chen, Qi and Doherty, Sarah J. and Freeman, Sarah and Forster, Piers M. and Fuglestvedt, Jan and Gettelman, Andrew and De Le{\'o}n, Rub{\'e}n Rodr{\'i}guez and Lim, Ling L. and Lund, Marianne T. and Millar, Richard J. and Owen, Bethan and Penner, Joyce E. and Pitari, Giovanni and Prather, Michael J. and Sausen, Robert and Wilcox, Laura J.},
  title   = {The contribution of global aviation to anthropogenic climate forcing for 2000 to 2018},
  journal = {Atmospheric Environment},
  volume  = {244},
  pages   = {117834},
  year    = {2021}
}

@article{karcher2018formation,
  author  = {K{\"a}rcher, Bernd},
  title   = {Formation and radiative forcing of contrail cirrus},
  journal = {Nature Communications},
  volume  = {9},
  pages   = {1824},
  year    = {2018}
}

@article{ng2024contrail,
  author  = {Ng, Joe Yue-Hei and McCloskey, Kevin and Cui, Jian and Meijer, Vincent R. and Brand, Erica and Sarna, Aaron and Goyal, Nita and Van Arsdale, Christopher and Geraedts, Scott},
  title   = {Contrail Detection on {GOES-16 ABI} With the {OpenContrails} Dataset},
  journal = {IEEE Transactions on Geoscience and Remote Sensing},
  volume  = {62},
  pages   = {1--14},
  year    = {2024}
}

@article{schmit2017abi,
  author  = {Schmit, Timothy J. and Griffith, Paul and Gunshor, Mathew M. and Daniels, Jaime M. and Goodman, Steven J. and Lebair, William J.},
  title   = {A Closer Look at the {ABI} on the {GOES-R} Series},
  journal = {Bulletin of the American Meteorological Society},
  volume  = {98},
  number  = {4},
  pages   = {681--698},
  year    = {2017}
}

@inproceedings{ho2020denoising,
  author    = {Ho, Jonathan and Jain, Ajay and Abbeel, Pieter},
  title     = {Denoising Diffusion Probabilistic Models},
  booktitle = {Advances in Neural Information Processing Systems},
  volume    = {33},
  year      = {2020}
}

@inproceedings{song2021denoising,
  author    = {Song, Jiaming and Meng, Chenlin and Ermon, Stefano},
  title     = {Denoising Diffusion Implicit Models},
  booktitle = {International Conference on Learning Representations},
  year      = {2021}
}

@article{amit2021segdiff,
  author  = {Amit, Tomer and Shaharbany, Tal and Nachmani, Eliya and Wolf, Lior},
  title   = {{SegDiff}: Image Segmentation with Diffusion Probabilistic Models},
  journal = {arXiv preprint arXiv:2112.00390},
  year    = {2021}
}

@inproceedings{wu2023medsegdiff,
  author    = {Wu, Junde and Fu, Rao and Fang, Huihui and Zhang, Yu and Yang, Yehui and Xiong, Haoyi and Liu, Huiying and Xu, Yanwu},
  title     = {{MedSegDiff}: Medical Image Segmentation with Diffusion Probabilistic Model},
  booktitle = {Medical Imaging with Deep Learning},
  year      = {2023}
}

@inproceedings{chen2018encoder,
  author    = {Chen, Liang-Chieh and Zhu, Yukun and Papandreou, George and Schroff, Florian and Adam, Hartwig},
  title     = {Encoder-Decoder with Atrous Separable Convolution for Semantic Image Segmentation},
  booktitle = {European Conference on Computer Vision},
  year      = {2018}
}

@inproceedings{he2016deep,
  author    = {He, Kaiming and Zhang, Xiangyu and Ren, Shaoqing and Sun, Jian},
  title     = {Deep Residual Learning for Image Recognition},
  booktitle = {IEEE Conference on Computer Vision and Pattern Recognition},
  year      = {2016}
}

@inproceedings{deng2009imagenet,
  author    = {Deng, Jia and Dong, Wei and Socher, Richard and Li, Li-Jia and Li, Kai and Fei-Fei, Li},
  title     = {{ImageNet}: A Large-Scale Hierarchical Image Database},
  booktitle = {IEEE Conference on Computer Vision and Pattern Recognition},
  year      = {2009}
}

@inproceedings{lin2017focal,
  author    = {Lin, Tsung-Yi and Goyal, Priya and Girshick, Ross and He, Kaiming and Doll{\'a}r, Piotr},
  title     = {Focal Loss for Dense Object Detection},
  booktitle = {IEEE International Conference on Computer Vision},
  year      = {2017}
}

@inproceedings{shit2021cldice,
  author    = {Shit, Suprosanna and Paetzold, Johannes C. and Sekuboyina, Anjany and Ezhov, Ivan and Unger, Alexander and Zhylka, Andrey and Pluim, Josien P. W. and Bauer, Ulrich and Menze, Bjoern H.},
  title     = {{clDice} -- A Novel Topology-Preserving Loss Function for Tubular Structure Segmentation},
  booktitle = {IEEE Conference on Computer Vision and Pattern Recognition},
  year      = {2021}
}

@article{ho2022classifier,
  author  = {Ho, Jonathan and Salimans, Tim},
  title   = {Classifier-Free Diffusion Guidance},
  journal = {arXiv preprint arXiv:2207.12598},
  year    = {2022}
}

@inproceedings{perez2018film,
  author    = {Perez, Ethan and Strub, Florian and de Vries, Harm and Dumoulin, Vincent and Courville, Aaron},
  title     = {{FiLM}: Visual Reasoning with a General Conditioning Layer},
  booktitle = {AAAI Conference on Artificial Intelligence},
  year      = {2018}
}

@inproceedings{wang2018esrgan,
  author    = {Wang, Xintao and Yu, Ke and Wu, Shixiang and Gu, Jinjin and Liu, Yihao and Dong, Chao and Qiao, Yu and Loy, Chen Change},
  title     = {{ESRGAN}: Enhanced Super-Resolution Generative Adversarial Networks},
  booktitle = {European Conference on Computer Vision Workshops},
  year      = {2018}
}

@article{teoh2020mitigating,
  author  = {Teoh, Roger and Schumann, Ulrich and Majumdar, Arnab and Stettler, Marc E. J.},
  title   = {Mitigating the Climate Forcing of Aircraft Contrails by Small-Scale Diversions and Technology Adoption},
  journal = {Environmental Science \& Technology},
  volume  = {54},
  number  = {5},
  pages   = {2941--2950},
  year    = {2020}
}

@misc{blueskies2026,
  author       = {{University of Cambridge}},
  title        = {Operation {Blue} {Skies} takes off: landmark trial to test {AI}-driven contrail avoidance},
  howpublished = {\url{https://www.cam.ac.uk/research/news/operation-blue-skies-takes-off-landmark-trial-to-test-ai-driven-contrail-avoidance}},
  year         = {2026},
  note         = {Accessed 19 August 2026}
}

@misc{googleblueskies2026,
  author       = {{Google}},
  title        = {Google and {UK} Government launch {AI} contrail avoidance trial},
  howpublished = {\url{https://blog.google/innovation-and-ai/models-and-research/google-research/blue-skies/}},
  year         = {2026},
  note         = {Accessed 19 August 2026}
}

@inproceedings{lakshminarayanan2017simple,
  author    = {Lakshminarayanan, Balaji and Pritzel, Alexander and Blundell, Charles},
  title     = {Simple and Scalable Predictive Uncertainty Estimation using Deep Ensembles},
  booktitle = {Advances in Neural Information Processing Systems},
  volume    = {30},
  year      = {2017}
}

@article{geraedts2024scalable,
  author  = {Geraedts, Scott and Brand, Erica and Dean, Thomas R. and Eastham, Sebastian and Elkin, Carl and Engberg, Zebediah and Hager, Ulrike and Langmore, Ian and McCloskey, Kevin and Ng, Joe Yue-Hei and Platt, John C. and Sankar, Tharun and Sarna, Aaron and Shapiro, Marc and Goyal, Nita},
  title   = {A scalable system to measure contrail formation on a per-flight basis},
  journal = {Environmental Research Communications},
  volume  = {6},
  number  = {1},
  pages   = {015008},
  year    = {2024}
}

@article{teoh2024global,
  author  = {Teoh, Roger and Engberg, Zebediah and Schumann, Ulrich and Voigt, Christiane and Shapiro, Marc and Rohs, Susanne and Stettler, Marc E. J.},
  title   = {Global aviation contrail climate effects from 2019 to 2021},
  journal = {Atmospheric Chemistry and Physics},
  volume  = {24},
  number  = {10},
  pages   = {6071--6093},
  year    = {2024}
}

@article{mannstein1999operational,
  author  = {Mannstein, Hermann and Meyer, Richard and Wendling, Peter},
  title   = {Operational detection of contrails from {NOAA-AVHRR}-data},
  journal = {International Journal of Remote Sensing},
  volume  = {20},
  number  = {8},
  pages   = {1641--1660},
  year    = {1999}
}

@article{vazquez2015contrail,
  author  = {V{\'a}zquez-Navarro, Margarita and Mannstein, Hermann and Kox, Stephan},
  title   = {Contrail life cycle and properties from 1 year of {MSG/SEVIRI} rapid-scan images},
  journal = {Atmospheric Chemistry and Physics},
  volume  = {15},
  number  = {15},
  pages   = {8739--8749},
  year    = {2015}
}

@mastersthesis{kulik2019satellite,
  author = {Kulik, Luke},
  title  = {Satellite-based detection of contrails using deep learning},
  school = {Massachusetts Institute of Technology},
  year   = {2019}
}

@article{meijer2022contrail,
  author  = {Meijer, Vincent R. and Kulik, Luke and Eastham, Sebastian D. and Allroggen, Florian and Speth, Raymond L. and Karaman, Sertac and Barrett, Steven R. H.},
  title   = {Contrail coverage over the {United States} before and during the {COVID-19} pandemic},
  journal = {Environmental Research Letters},
  volume  = {17},
  number  = {3},
  pages   = {034039},
  year    = {2022}
}

@misc{kaggle2023contrails,
  author       = {{Google Research}},
  title        = {Google {Research} -- {Identify} {Contrails} to {Reduce} {Global} {Warming}},
  howpublished = {Kaggle competition, \url{https://www.kaggle.com/competitions/google-research-identify-contrails-reduce-global-warming}},
  year         = {2023}
}

@article{chevallier2023linear,
  author  = {Chevallier, R{\'e}mi and Shapiro, Marc and Engberg, Zebediah and Soler, Manuel and Delahaye, Daniel},
  title   = {Linear contrails detection, tracking and matching with aircraft using geostationary satellite and air traffic data},
  journal = {Aerospace},
  volume  = {10},
  number  = {7},
  pages   = {578},
  year    = {2023}
}

@inproceedings{mosinska2018beyond,
  author    = {Mosinska, Agata and M{\'a}rquez-Neila, Pablo and Kozi{\'n}ski, Mateusz and Fua, Pascal},
  title     = {Beyond the Pixel-Wise Loss for Topology-Aware Delineation},
  booktitle = {IEEE Conference on Computer Vision and Pattern Recognition},
  year      = {2018}
}

@inproceedings{hu2019topology,
  author    = {Hu, Xiaoling and Li, Fuxin and Samaras, Dimitris and Chen, Chao},
  title     = {Topology-Preserving Deep Image Segmentation},
  booktitle = {Advances in Neural Information Processing Systems},
  volume    = {32},
  year      = {2019}
}

@inproceedings{lin2020deep,
  author    = {Lin, Yancong and Pintea, Silvia L. and van Gemert, Jan C.},
  title     = {Deep {Hough}-Transform Line Priors},
  booktitle = {European Conference on Computer Vision},
  year      = {2020}
}

@article{zhao2022deep,
  author  = {Zhao, Kai and Han, Qi and Zhang, Chang-Bin and Xu, Jun and Cheng, Ming-Ming},
  title   = {Deep {Hough} Transform for Semantic Line Detection},
  journal = {IEEE Transactions on Pattern Analysis and Machine Intelligence},
  volume  = {44},
  number  = {9},
  pages   = {4793--4806},
  year    = {2022}
}

@inproceedings{wolleb2022diffusion,
  author    = {Wolleb, Julia and Sandk{\"u}hler, Robin and Bieder, Florentin and Valmaggia, Philippe and Cattin, Philippe C.},
  title     = {Diffusion Models for Implicit Image Segmentation Ensembles},
  booktitle = {Medical Imaging with Deep Learning},
  year      = {2022}
}
}

\newpage
\appendix
\section{Reproducibility details}
\label{app:repro}
\paragraph{Code.}
\url{https://github.com/spandan1305/contrail_detection}
\paragraph{Training.}
All diffusion variants train for 90 epochs with AdamW (learning rate $10^{-4}$,
cosine schedule, batch 32) on a single GPU; inference uses 10 DDIM steps and
$N{=}20$ chains. Tables report seed 42; a second seed differs by $0.002$
PR-AUC (two-seed mean $0.475\pm0.001$). The same 50-pixel minimum-component
filter is applied to every model's binary mask, and operating thresholds are
calibrated per model on val-tune only (vote 0.6 for ours, 0.5 for the others).

\paragraph{Baseline adaptation.}
MedSegDiff is the official implementation adapted to this dataset
(MedSegDiff-B recipe). Of its two output heads, the calibration head was
selected on val-tune and only that head was scored on test. It keeps its own
single-frame conditioning and loss stack, which is its published design. A
control rerun conditioned on all four frames under the same recipe and budget
scored \emph{worse} ($0.033$ against $0.119$ test PR-AUC), so the table
reports the stronger variant.

\paragraph{The gap to the published OpenContrails model.}
The external reference row ($0.651$) is not a within-protocol comparison: the
model was trained under a much larger budget and evaluated on the full
1{,}856-scene pool. The gap is not a labeling artifact, because per-scene
positive-pixel counts for all 1{,}856 raw validation masks match our
preprocessed masks exactly. It is also not a prevalence artifact: measured
prevalence is $0.181\%$ over the pool and $0.184\%$ over our test half, so the
often-quoted $1.2\%$ figure describes their contrail-enriched training set,
not the evaluation pool.

\end{document}